\documentclass[10pt,twocolumn,letterpaper]{article}

\usepackage{booktabs} 
\usepackage{cvpr}
\usepackage{times}
\usepackage{epsfig}
\usepackage{graphicx}
\usepackage{amsmath}
\usepackage{amssymb}
\usepackage{upgreek}
\usepackage{siunitx}

\usepackage[breaklinks=true,bookmarks=false]{hyperref}

\cvprfinalcopy 

\ifcvprfinal\fi

\begin{document}
	
\title{%
Polka Lines: \\Learning Structured Illumination and Reconstruction for Active Stereo
} 
\author{
 \hspace{-10mm}
Seung-Hwan Baek
\qquad
Felix Heide  \vspace{2mm}\\
	Princeton University
}



\twocolumn[{%
\vspace{-7mm}
\renewcommand\twocolumn[1][]{#1}%
\maketitle
\vspace{-3mm}
\thispagestyle{empty}
}]
\begin{abstract}
Active stereo cameras that recover depth from structured light captures have become a cornerstone sensor modality for 3D scene reconstruction and understanding tasks across application domains.
Active stereo cameras project a pseudo-random dot pattern on object surfaces to extract disparity independently of object texture.
Such hand-crafted patterns are designed in isolation from the scene statistics, ambient illumination conditions, and the reconstruction method.
In this work, we propose a method to jointly learn structured illumination and reconstruction, parameterized by a diffractive optical element and a neural network, in an end-to-end fashion.
To this end, we introduce a differentiable image formation model for active stereo, relying on both wave and geometric optics, and a trinocular reconstruction network.
The jointly optimized pattern, which we dub ``Polka Lines,'' together with the reconstruction network, makes accurate active-stereo depth estimates across imaging conditions.
We validate the proposed method in simulation and using with an experimental prototype, and we demonstrate several variants of the Polka Lines patterns specialized to the illumination conditions.
\end{abstract}

\vspace{-6mm}
\section{Introduction}
\label{sec:introduction}
Active depth cameras have become essential for three-dimensional scene reconstruction and scene understanding, with established and emerging applications across disciplines, including robotics, autonomous drones, navigation, driver monitoring, human-computer interaction, virtual and mixed reality, and remote conferencing. When combined with RGB cameras, depth-sensing methods have made it possible to recover high-fidelity scene reconstructions~\cite{izadi2011kinectfusion}. Such RGB-D cameras also allowed researchers to collect large-scale RGB-D data sets that propelled work on fundamental computer vision problems, including scene understanding~\cite{song2015sun,hickson2014efficient} and action recognition~\cite{ni2013rgbd}. However, while depth cameras under controlled conditions with low ambient light and little object motion are becoming reliable~\cite{intelD415,sell2014xbox}, depth imaging in strong ambient light, at long ranges, and for fine detail and highly dynamic scenes remains an open challenge.

A large body of work has explored active depth sensing approaches to tackle this challenge~\cite{hansard2012time,lange00tof,achar2017epipolar,scharstein2003high}, with structure light and time-of-flight cameras being the most successful methods. Pulsed time-of-flight sensors emit pulses of light into the scene and measure the travel time of the returned photons directly by employing sensitive silicon avalanche photo-diodes~\cite{williams_apds} or single-photon avalanche diodes~\cite{aull2002geiger}. Although these detectors are sensitive to a single photon, their low fill factor restricts existing LiDAR sensors to point-by-point scanning with individual diodes, which prohibits the acquisition of dense depth maps. Correlation time-of-flight sensors \cite{hansard2012time,kolb2010time,lange00tof} overcome this challenge by indirectly estimating round-trip time from the phase of temporally modulated illumination. Although these cameras provide accurate depth for indoor scenes, they suffer from strong ambient illumination and multi-path interference~\cite{su2018deep, marco2017deeptof}, are limited to VGA resolution, and they require multiple captures, which makes dynamic scenes a challenge. Active stereo~\cite{zhang2018activestereonet,intelD415,google_pix4_depth} has emerged as the only low-cost depth sensing modality that has the potential to overcome these limitations of existing methods for room-sized scenes. Active stereo cameras equip a stereo camera pair with an illumination module that projects a fixed pattern onto a scene so that, independently of surface texture, stereo correspondence can be reliably estimated. As such, active stereo methods allow for single-shot depth estimates at high resolutions using low-cost diffractive laser dot modules~\cite{intelD415} and conventional CMOS sensor deployed in mass-market products including Intel RealSense cameras~\cite{intelD415} and the Google Pixel 4 Phones~\cite{google_pix4_depth}.
However, although active stereo has become a rapidly emerging depth-sensing technology, existing approaches struggle with extreme ambient illumination and complex scenes, prohibiting reliable depth estimates in uncontrolled in-the-wild scenarios.

These limitations are direct consequences of the pipeline design of existing active stereo systems, which hand-engineer the illumination patterns and the reconstruction algorithms in isolation. Typically, the illumination pattern is designed in a first step using a diffractive optical element (DOE) placed in front of a laser diode. Existing dot patterns resulting from known diffractive gratings, such as the Dammann grating~\cite{dammann1971high}, are employed with the assumption that generating uniform textures ensures robust disparity estimation for the average scene.
Given a fixed illumination pattern, the reconstruction algorithm is then designed with the goal of estimating correspondence using cost-volume methods~\cite{bleyer2011patchmatch,hirschmuller2007stereo} or learning-based methods~\cite{ryan2016hyperdepth,fanello2017ultrastereo,zhang2018activestereonet,riegler2019connecting}.
In this conventional design paradigm, the illumination pattern does not receive feedback from the reconstruction algorithm or the dataset of scenes, prohibiting end-to-end learning of optimal patterns, reconstruction algorithms, and capture configurations tailored to the scene.

In this work, we propose a method that jointly learns illumination patterns and a reconstruction algorithm, parameterized by a DOE and a neural network, in an end-to-end manner.
The resulting optimal illumination patterns, which we dub ``Polka Lines'', together with the reconstruction network, allow for high-quality scene reconstructions.
Moreover, our method allows us, for the first time, to learn environment-specific illumination patterns for active stereo systems. 
The proposed method hinges on a differentiable image formation model that relies on wave and geometric optics to make the illumination and capture simulation accurate and, at the same time, efficient enough for joint optimization. 
{We then propose a trinocular active stereo network that estimates an accurate depth map from the sensor inputs. Unlike previous methods that only use binocular inputs from the stereo cameras, our network exploits the known illumination pattern, resulting in a trinocular stereo setup which reduces reconstruction errors near occlusion boundaries}.
We train the fully differentiable illumination and reconstruction model in a supervised manner and finetune the reconstruction for an experimental prototype in a self-supervised manner.
The proposed Polka Lines patterns, together with the reconstruction network, allows us to achieve state-of-the-art active stereo depth estimates for a wide variety of imaging conditions.
%

Specifically, We make the following contributions:
\vspace{-3pt}
\begin{itemize}
	\setlength\itemsep{.2em}
  \item We introduce a novel differentiable image formation model for active stereo systems based on geometric and wave optics.
    \item We devise a novel trinocular active stereo network that uses the known illumination pattern in addition to the stereo inputs. 
	\item We jointly learn optimal ``Polka Lines'' illumination patterns via differentiable end-to-end optimization, which can be specialized to specific illumination conditions.
  \item We validate the proposed method in simulation and with an experimental prototype. We demonstrate robust depth acquisition across diverse scene scenarios from low light to strong illumination.
\end{itemize} 
\section{Related Work}
\label{sec:relatedwork}

\vspace{0.5em}\noindent\textbf{Depth Imaging.}\hspace{0.1em}
Depth cameras can be broadly categorized into two families, passive and active cameras. 
Passive methods exploit depth cues such as parallax~\cite{scharstein2002taxonomy,godard2017unsupervised}, defocus~\cite{levin2007image}, and double refraction~\cite{baek2016birefractive,Meuleman_2020_CVPR} that do not require illumination control.
Passive methods often fail on challenging scene parts, such as textureless surfaces, where they can produce catastrophic depth estimation errors.
Active systems employ specialized illumination modules to tackle textureless surfaces. Major directions include pulsed and continuous-wave time-of-flight sensors~\cite{heide2015doppler,heide2018sub}, gated imaging~\cite{gruber2019gated2depth}, structured-light sensor~\cite{gupta2013structured,wu2016snapshot}, and active stereo systems~\cite{zhang2018activestereonet}.
Among these, active stereo is particularly attractive as it promises robust single-shot depth imaging at low system cost and small form factor. As such, active stereo systems have successfully been deployed in mass-market~\cite{intelD415,google_pix4_depth}.
However, existing active-stereo systems also struggle in challenging environments with strong ambient light and noisy inputs with varying scene reflectance.
This reduced accuracy partly originates from the blind, compartmentalized design process of the illumination pattern, which often does not consider the reconstruction method, scene statistics, and illumination conditions.
In this work, we close this gap by proposing to jointly optimize the illumination patterns and the reconstruction method for active stereo.

%

\vspace{0.5em}\noindent\textbf{Illumination Patterns for Active Stereo.}\hspace{0.1em}
Designing an illumination pattern is crucial for the accuracy of correspondence matching in active stereo systems.
Existing methods commonly employ Dammann gratings~\cite{dammann1971high} and Vertical Cavity Surface Emitting Lasers that result in locally-distinct, but globally repetitive illumination patterns~\cite{martinez2013kinect,kowdle2018need,intelD415}.
This heuristic design is blind to scene statistics, noise levels, and the reconstruction method.
Existing methods have attempted to improve depth estimation by employing alternative hand-crafted DOE designs~\cite{du2016design,vandenhouten2017design,miao2019design} that rely on alternative experts and heuristic metrics on the illumination patterns.
We depart from these heuristic designs and instead directly optimize the illumination pattern with the depth reconstruction accuracy as a loss via end-to-end optimization.

\vspace{0.5em}\noindent\textbf{Active Stereo Depth Estimation.}\hspace{0.1em}
Depth reconstruction for active-stereo systems aims to estimate accurate correspondence between stereo images with the aid of projected illumination patterns for feature matching.
The corresponding large body of work can be categorized into methods relying on classic patch-based correspondence matching~\cite{hirschmuller2007stereo,bleyer2011patchmatch} and recent learning-based methods~\cite{ryan2016hyperdepth,fanello2017ultrastereo,zhang2018activestereonet,riegler2019connecting}.
Zhang et al.~\cite{zhang2018activestereonet} proposed an active stereo network with self-supervision, removing the cumbersome process of acquiring training data, and improving depth estimation accuracy.
All of these existing reconstruction methods are limited by the fixed illumination pattern. As such, these methods have to adapt to a given pattern and cannot vary the pattern to suit different imaging conditions. We jointly optimize the illumination and reconstruction module, allowing us to tailor the pattern to the reconstruction method and scene statistics.
Moreover, departing from existing approaches, the proposed trinocular reconstruction is the first that exploits knowing illumination pattern itself.

\vspace{0.5em}\noindent\textbf{Differentiable Optics.}\hspace{0.1em}
With the advent of auto-differentiation frameworks~\cite{tensorflow2015-whitepaper,paszke2017automatic}, jointly optimizing imaging optics and reconstruction methods has shaped the design process of diverse vision systems~\cite{chakrabarti2016learning,lizhi8552450,nehme20deepstorm3d,sun2020endtoendspad,haim2018depth,wu2019phasecam3d,Chang:2019:DeepOptics3D,sitzmann2018end,Metzler:2020:DeepOpticsHDR,Sun2020LearningRank1HDR}.
While existing methods have focused on the imaging optics and primarily assume near-field propagation, we instead optimize illumination optics, specifically a DOE in front of a collimated laser, using far-field wave propagation from a laser to the scene. At the same time, we rely on ray optics to simulate stereo imaging via epipolar geometry.
This hybrid image formation, which exploits both wave and geometric optics, allows us to efficiently simulate light transport in active stereo systems while being efficient enough for gradient-based end-to-end optimization.
We note that Wu et al.~\cite{wu2020freecam3d} proposed a depth-from-defocus method with a learned aperture mask for structured-light systems.
However, this blur-based structured-light projection suffers from frequency-limited features. As such, it is orthogonal to the proposed method, which optimizes a diffraction pattern at the far field for active stereo.
Related optimization principles for illumination design can also be found in reflectance imaging~\cite{kang2019learning}.

\section{Differentiable Hybrid Image Formation}
\label{sec:optics}

\setlength{\belowdisplayskip}{5pt} \setlength{\belowdisplayshortskip}{5pt}
\setlength{\abovedisplayskip}{5pt} \setlength{\abovedisplayshortskip}{5pt}

\begin{figure}[t]
  \centering
  \includegraphics[width=\linewidth]{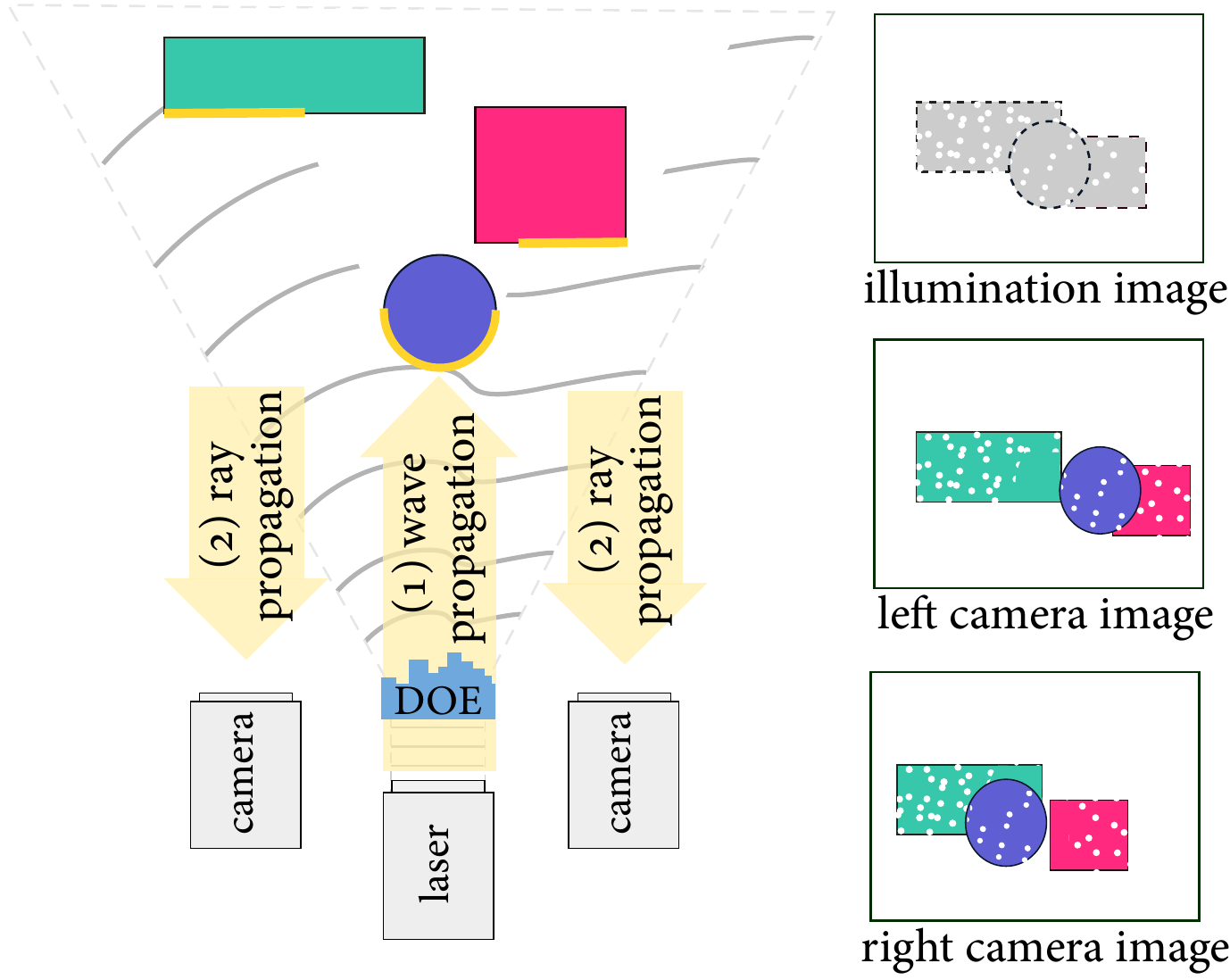}
  \caption{\label{fig:diagram}
  We simulate the illumination image projected by the laser and the DOE using wave optics.
  We then simulate the stereo images captured by cameras using geometric optics.
  \vspace{-4mm}
  }
\end{figure}

To jointly learn structured illumination patterns and reconstruction methods, we introduce a differentiable image formation model for active stereo sensing.
Active stereo systems consist of stereo cameras and an illumination module that codes light with a laser-illuminated DOE as shown in Figure~\ref{fig:diagram}.
The light transport of an active stereo system can be divided into two parts: one describing the propagation of the laser light into the scene with the output of the illumination pattern cast onto the scene, and the other describing the illumination returned from the scene to the stereo cameras.
We rely on wave optics for the former part and geometric optics for the latter part, comprising the proposed hybrid image formation model.

\subsection{Modeling the Projected Illumination Pattern}
Simulating light transport from an active stereo illumination module to a scene amounts to computing the illumination pattern projected onto the scene from the laser (Figure~\ref{fig:diagram}).
Relying on wave optics, we represent the light emitted by the laser as amplitude $A$ and phase $\phi$ at each discrete spatial location $x, y$ sampled with pitch $u$ and with $N\times N$ resolution\footnote{$u=\SI{1}{\micro m}$ and $N=1000$ in our experiments.}.

\vspace{0.5em}\noindent\textbf{Phase Delay on the DOE.}\hspace{0.1em}
The phase of the emitted light wave is modulated when it passes through the DOE by $\phi_{\text{delay}}$ as $\phi \leftarrow \phi+\phi_{\text{delay}}$.
The phase delay $\phi_{\text{delay}}$ is related to the height of the DOE $h$, the wavelength of the light $\lambda$, and the refractive index of the DOE for that wavelength $\eta_{\lambda}$, that is
\begin{equation}\label{eq:phase_delay_doe}
\phi_{\text{delay}} = \frac{2 \pi {(\eta_\lambda-1)} }{\lambda} h.
\end{equation}

\vspace{0.5em}\noindent\textbf{Far-field Wave Propagation.}\hspace{0.1em}
Next, the light wave modulated by the DOE propagates into the scene.
We model this propagation using Fraunhofer far-field wave propagation because we assume that scene depth ranges from $\SI{0.4}{m}$ to $\SI{3}{m}$ which is sufficiently larger than the wave spatial extent $uN = \SI{1}{mm}$~\cite{goodman2005introduction}.
We implement this propagation operation by computing the Fourier transform $\mathcal{F}$ of the complex-valued light wave $U$ of amplitude $A$ and phase $\phi$
\begin{equation}\label{eq:fraunhofer}
  U' \leftarrow \mathcal{F}(U),
\end{equation}
where $U'$ is the propagated complex light wave.
Finally, the illumination pattern $P$ in the scene is the intensity of the propagated light wave, a squared magnitude of $U'$
\begin{equation}
P \leftarrow |U'|^2.
\end{equation}
The resolution of the pattern $P$ remains the same as that of $U$, while the physical pixel pitch $v$ of the pattern $P$ changes accordingly as $v=\frac{\lambda z} {uN}$, where $z$ is the propagation distance~\cite{goodman2005introduction}.
Refer to the Supplemental Document for the simulated illumination patterns corresponding to existing DOE designs.

\vspace{0.5em}\noindent\textbf{Sampling the Illumination Pattern.}\hspace{0.1em}
A pixel in the simulated illumination image $P$ has the physical width of $v=\frac{\lambda z}{uN}$ at a scene depth $z$.
At the same time, a camera pixel maps to a width of $\frac{p}{f}z$ at the scene depth $z$ via perspective unprojection, where $f$ is the camera focal length, and $p$ is the pixel pitch of the camera.
We resample the illumination image $P$ to have the same pixel pitch as a camera pixel pitch.
We compute the corresponding scale factor as follows
\begin{equation}\label{eq:scale}
  \frac{\mathrm{camera \; pixel \; size}}{\mathrm{illumination \; pattern \; pixel \; size}}=\frac{\frac{p}{f}z}{\frac{\lambda} {uN}z} = \frac{puN}{f\lambda}.
\end{equation}
The scale factor $\frac{puN}{f\lambda}$ is applied to the illumination image
 $P \leftarrow \textsf{resample}(P, \frac{puN}{f\lambda})$,
where $\textsf{resample}$ is the bicubic resampling operator.

Note that the depth dependency for the pixel sizes for the illumination pattern and the camera disappears in the scaling factor, meaning that the scale factor is independent of the propagation distance of the light.
This indicates that the illumination pattern $P$ can be applied to any scene regardless of its depth composition, which facilitates efficient simulation of the light transport.

\subsection{Synthesis of Stereo Images}
Once the illumination image $P$ is computed, we then simulate stereo images. 
While wave optics can describe this procedure using Wigner distribution functions and far-field wave propagation, this would be prohibitively expensive for the proposed end-to-end optimization procedure, which requires tens of thousands of iterations, each triggering multiple forward simulations.
Instead, we use a geometric-optics model representing light using intensity only, instead of both phase and amplitude as in wave optics.

\vspace{0.5em}\noindent\textbf{Light-matter Interaction and Measurement.}\hspace{0.1em}
Given the illumination image $P$ at the viewpoint of the illumination module, we next simulate the light-matter interaction and sensor measurement by the stereo cameras.
In the following model, we use disparity maps $D^\text{L/R}$, reflectance maps $I^\text{L/R}$, and occlusion masks $O^\text{L/R}$ at the left and the right camera viewpoints.
Occlusion masks $O^\text{L/R}$ describe the visibility at the viewpoints of the left/right camera with respect to the illumination module.

We first warp the illumination image $P$ to the left and the right camera viewpoints using the disparity $D^\text{L/R}$. We incorporate the occlusion maps $O^\text{L/R}$ through element-wise multiplication with the warped images, resulting in the final illumination images seen at the stereo camera viewpoints ($P^\text{L}$ and $P^\text{R}$), that is,
\begin{equation}\label{eq:stereo_illumination}
  P^{\text{L/R}} = O^\text{L/R} \odot  \textsf{warp}(P, D^\text{L/R}),
\end{equation}
where $\odot$ is the element-wise product and the operator $\textsf{warp}$ warps the illumination image $P$ by the disparity $D^\text{L/R}$.

We then compute scene response and sensor measurement using a Lambertian reflectance model. We implement imaging parameters including sensor clipping, signal-independent Gaussian noise, camera exposure, illumination power, and ambient illumination. Altogether, this is described by
\begin{equation}\label{eq:image_formation}
  J^{\text{L/R}} = \sigma ( \gamma (\alpha+ \beta P^{\text{L/R}})I^{\text{L/R}} + \eta),
\end{equation}
where $J^\text{L/R}$ are the simulated captured images for the left and the right camera viewpoints. The term
$\gamma$ is the scalar describing exposure and the sensor's spectral quantum efficiency,
$\alpha$ is the ambient light,
$\beta$ is the power of the laser illumination,
$\eta$ is Gaussian noise,
and $\sigma$ is the intensity-cropping function.


\begin{figure}[t]
  \centering
  \includegraphics[width=0.95\linewidth]{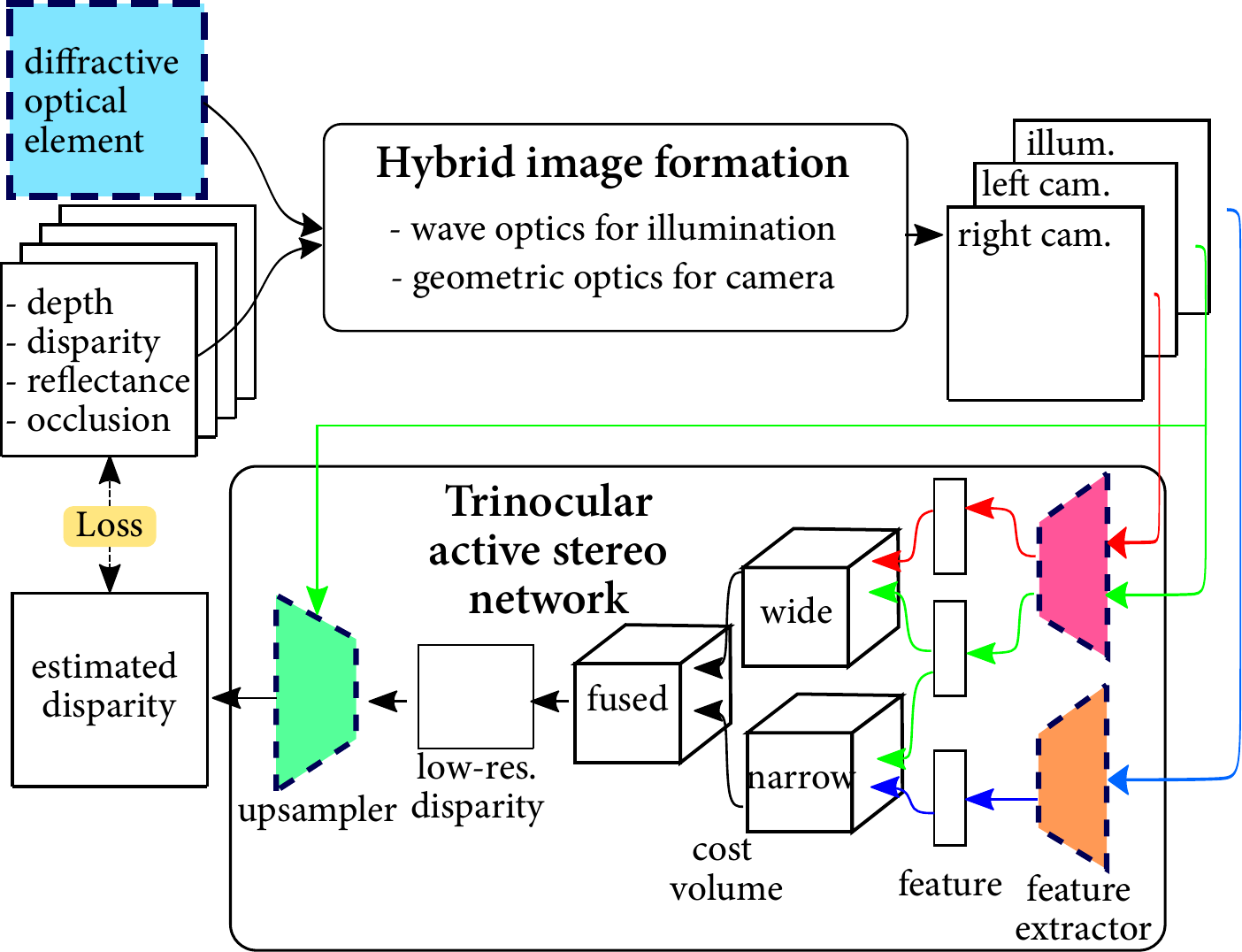}
  \caption{\label{fig:network}
  The proposed hybrid image formation model simulates the stereo images from which we reconstruct a depth map using a trinocular network. The loss is backpropagated to both the DOE and the network, enabling joint optimization. Dotted boxes indicate optimization parameters.
  \vspace{-5mm}
  }
\end{figure}
\section{Trinocular Active Stereo Network}
\label{sec:reconstruction}
%
We depart from existing active stereo architectures that take stereo images or a single illumination image as inputs~\cite{zhang2018activestereonet,riegler2019connecting}. Instead, we exploit the fact that an active stereo system provides stereo cues between the cameras but also the illumination and camera pairs. Specifically, we consider two baseline configurations in our active stereo camera: a narrow-baseline configuration between the illumination module and either of the two cameras, and one wide-baseline pair consisting of the left and right cameras.
To take advantage of these two different baselines, we propose the following trinocular active stereo network, which is illustrated in Figure~\ref{fig:network}.

\vspace{0.5em}\noindent\textbf{Reconstruction Network.}\hspace{0.1em}
The proposed reconstruction network receives the following inputs: a left-camera image $\mathbf{x}_\text{L}$, a right-camera image $\mathbf{x}_\text{R}$, and an illumination image $\mathbf{x}_\text{illum}$.
During the training phase, our image formation model synthetically generates these trinocular inputs; during real-world testing, we directly use the calibrated sensor inputs.

\begin{figure}[t]
  \centering
  \includegraphics[width=\linewidth]{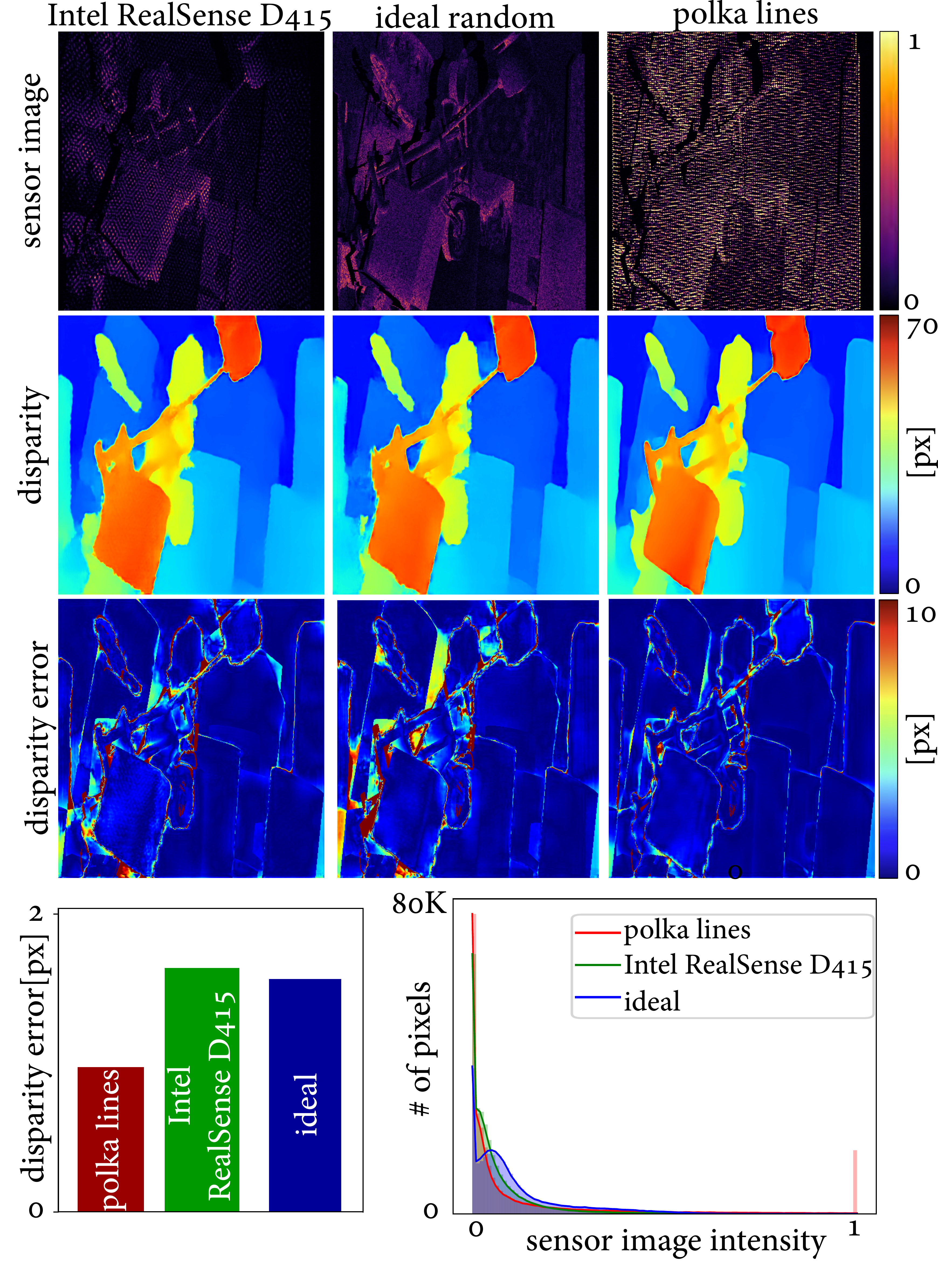}
  \caption{\label{fig:illumination}
   We evaluate our learned illumination pattern in simulation and we outperform the hand-crafted illumination pattern (Intel RealSense D415) and the ideal random pattern. Our learned Polka Line pattern effectively focuses energy to promote feature matching. The example shown here features an indoor environment.
   }
  \vspace{-6.5mm}
\end{figure}
The proposed network first extracts feature tensors $\mathbf{y}_\text{L/R/illum}$ of the three input images using two convolutional encoders: $\text{FE}_\text{cam}$ for the camera images and $\text{FE}_\text{illum}$ for the illumination image, that is
\begin{equation}\label{eq:feature_extraction}
\begin{array}{l}
{{\mathbf{y}}_{{\mathrm{L}}}} = {\mathrm{F}}{{\mathrm{E}}_{{\mathrm{cam}}}}({{\mathbf{x}}_{{\mathrm{L}}}}),
{{\mathbf{y}}_{{\mathrm{R}}}} = {\mathrm{F}}{{\mathrm{E}}_{{\mathrm{cam}}}}({{\mathbf{x}}_{{\mathrm{R}}}}),\\
{{\mathbf{y}}_{{\mathrm{illum}}}} = {\mathrm{F}}{{\mathrm{E}}_{{\mathrm{illum}}}}({{\mathbf{x}}_{{\mathrm{illum}}}}).
\end{array}
\end{equation}
Next, we construct trinocular cost volumes for two separate baselines.
We define a feature cost volume $C_\text{wide}$ for the wide-baseline pair as
\begin{equation}\label{eq:cost_volume_wide}
 C^d_\mathrm{wide}(x,y)=\mathbf{y}_\mathrm{L}(x,y) - \mathbf{y}_\mathrm{R}(x-d,y),
\end{equation}
where $d$ is a disparity candidate.
Similarly, the narrow-baseline cost volume is defined between the left-camera features $\mathbf{y}_\text{L}$ and the illumination features $\mathbf{y}_\text{illum}$ as
\begin{equation}\label{eq:cost_volume_narrow}
 C^d_\mathrm{narrow}(x,y)=\mathbf{y}_\mathrm{L}(x,y) - \mathbf{y}_\mathrm{illum}(x-d,y).
\end{equation}
We fuse the two cost volumes into a single cost volume
\begin{equation}\label{eq:cost_volume_fused}
  C^d_\mathrm{fused} = C^d_\mathrm{wide} + C^{\hat{d}}_\mathrm{narrow},
\end{equation}
where $\hat{d}=d\frac{b_\text{wide}}{b_\text{narrow}}$ is the disparity scaled by the ratio between the wide baseline and the narrow baseline.
Per-pixel disparity probability is computed using a soft-max layer, followed by disparity regression on the obtained probability resulting from the low-resolution disparity estimate~\cite{zhang2018activestereonet}.
Finally, an edge-aware convolutional upsampler estimates a disparity map $D^\text{L}_\text{est}$ for the left camera viewpoint at the original resolution.
For network details, we refer the reader to the Supplemental Document.

\begin{figure}[t]
  \centering
  \includegraphics[width=0.9\linewidth]{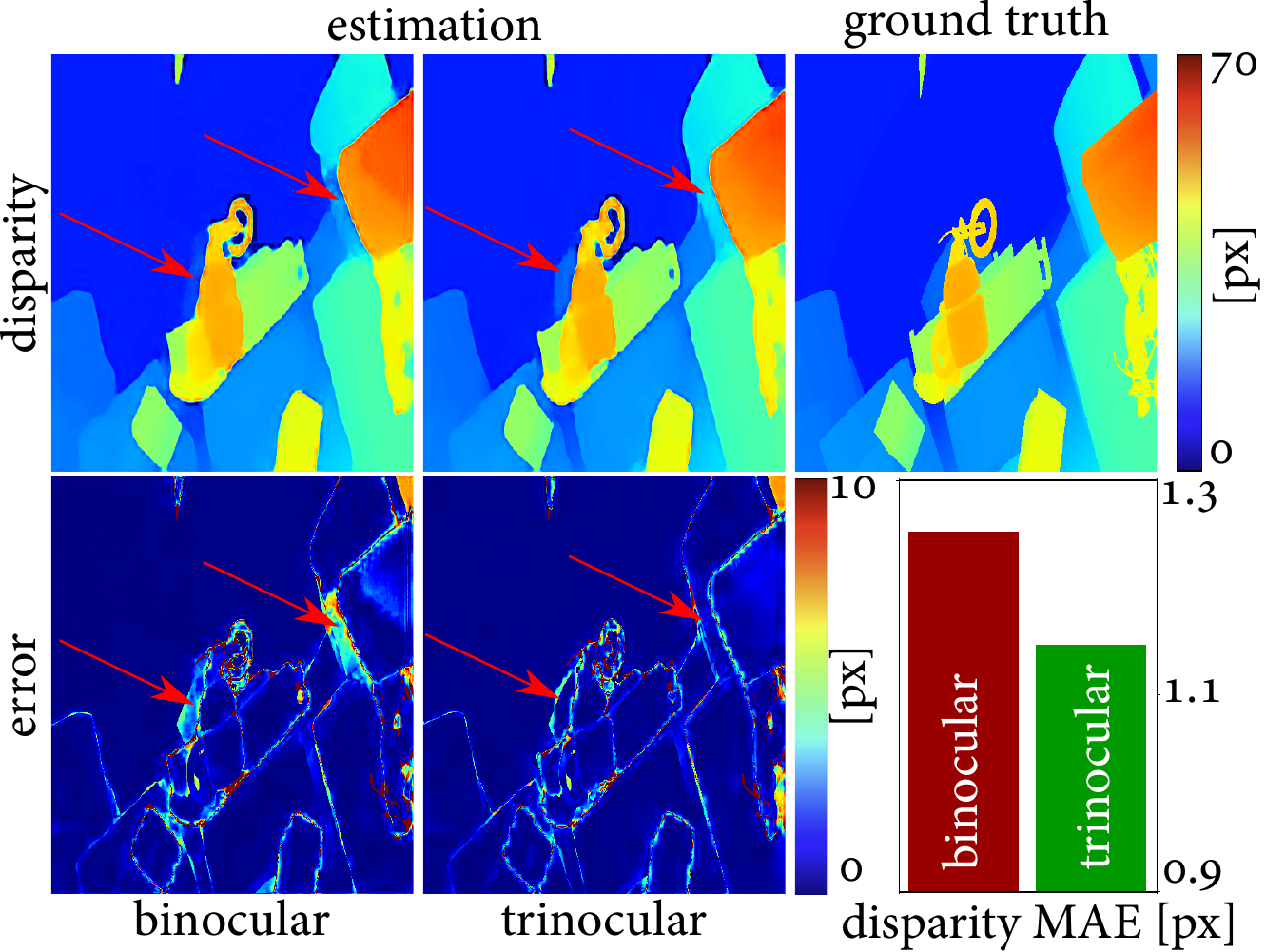}
  \caption{\label{fig:trinocular}
  The proposed trinocular reconstruction approach is more robust at object boundaries than conventional binocular methods, as it exploits cues between several camera and illumination pairs in a single active stereo system.
  \vspace{-6mm}
  }
\end{figure}

\vspace{0.5em}\noindent\textbf{Joint Learning.}\hspace{0.1em}
Denoting the network parameters as $\theta$ and the phase delay for the DOE as $\phi_\mathrm{delay}$, we solve the following end-to-end joint optimization problem
\begin{equation}\label{eq:opt}
\begin{array}{l}
  \mathop {{\rm{minimize}}}\limits_{\phi_\mathrm{delay} ,\theta } \mathcal{L}_{\text{s}}(D^\text{L}_\text{est}\left(\phi_\mathrm{delay}, \theta\right), D^\text{L}),
\end{array}
\end{equation}
where $\mathcal{L}_{\text{s}} = \textsf{MAE}$ is the mean-absolute-error loss of the estimated disparity supervised by the ground-truth disparity $D^\text{L}$.
Note that solving this optimization problem using stochastic gradient methods is only made possible by formulating the proposed image formation model and reconstruction method as fully differentiable operations.
We also incorporate varying ambient illumination conditions into our learning framework by controlling the following simulation parameters: ambient light power $\alpha$ and scalar $\gamma$ in Equation~\eqref{eq:image_formation}.
We train three separate models for different illumination configurations of generic, indoor, and outdoor environments.
For details, we refer the reader to the Supplemental Document.

\vspace{0.5em}\noindent\textbf{Dataset.}\hspace{0.1em}
Our method requires an active-stereo dataset of disparity maps $D^\text{L/R}$, NIR reflectance maps $I^\text{L/R}$, and occlusion masks $O^\text{L/R}$ at the left and the right camera viewpoints.
To obtain this dataset, we modify a synthetic passive-stereo RGB dataset~\cite{MIFDB16} which provides disparity maps $D^\text{L/R}$  but not the NIR reflectance maps $I^\text{L/R}$ and the occlusion masks $O^\text{L/R}$.
We obtain the NIR reflectance maps $I^\text{L/R}$ from the RGB stereo images using the RGB-inversion method from \cite{gruber2019gated2depth}.
Next, we compute the occlusion masks $O^\text{L/R}$ of the stereo cameras with respect to the illumination module.
We horizontally shrink the stereo occlusion masks by half since the illumination module lies halfway between the stereo pair.
Finally, we resize the images to the same resolution as the illumination images.
%
\begin{figure}[t]
\vspace{-3mm}
  \centering
  \includegraphics[width=0.9\linewidth]{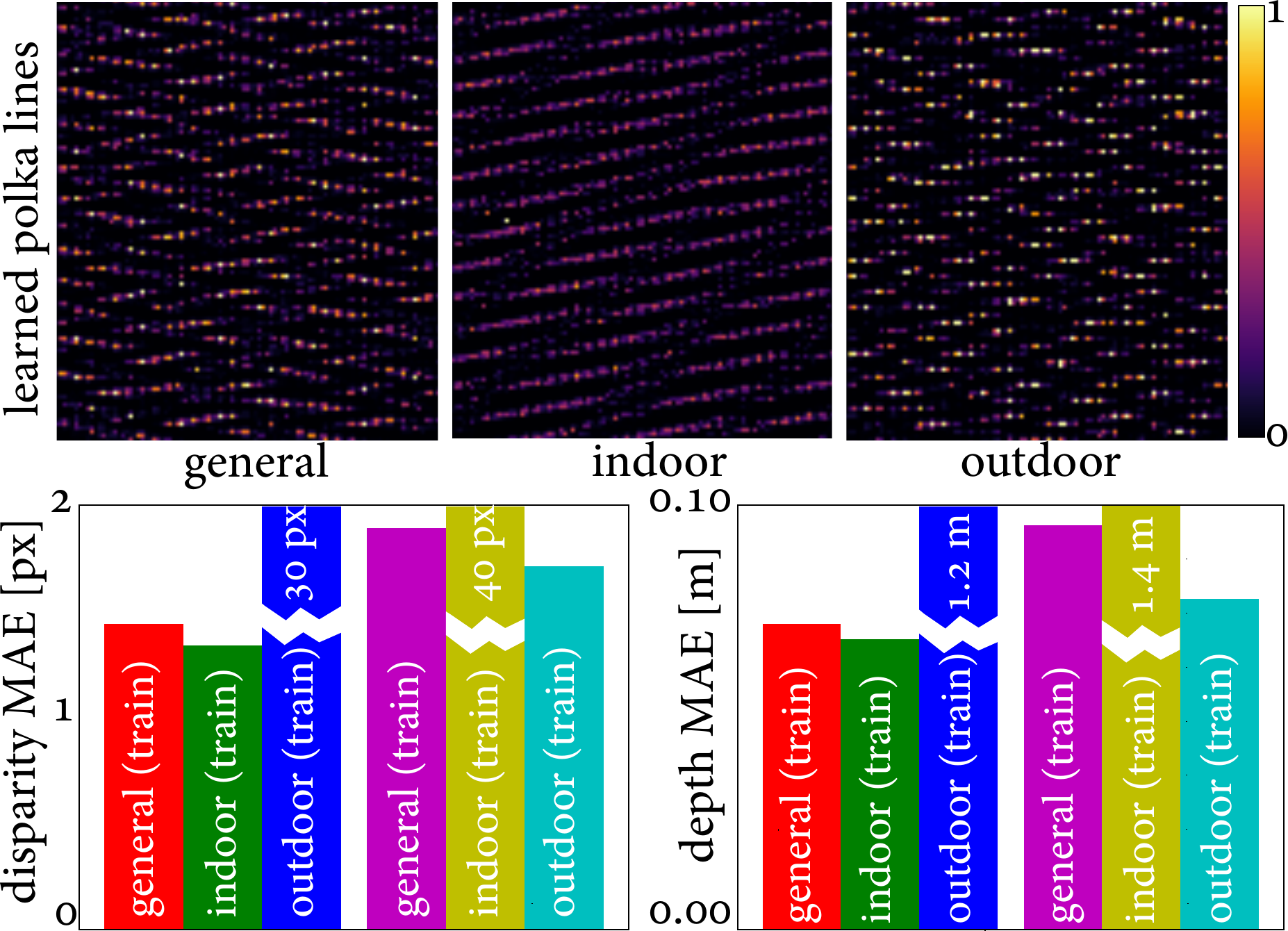}
  \caption{\label{fig:ambient}
  By changing simulation parameters, the proposed end-to-end optimization method can learn illumination patterns tailored to indoor, outdoor, and generic environments.
  \vspace{-4mm}
  }
\end{figure}
\vspace{-2mm}
\section{Self-supervised Finetuning}
To compensate for fabrication inaccuracies of the optimized DOE and the domain gap between the simulated training images and the real captures, we finetune our reconstruction network using a real-world dataset captured by our prototype.
To this end, we capture left and right IR image pairs $J^\text{L/R}$ and obtain the illumination images $P^\text{L/R}$ by projecting patterns onto a diffuse textureless wall.
However, for the disparity maps and the occlusion masks, it is challenging to obtain corresponding ground truths in the real world.
Therefore, we adopt the self-supervised learning approach previously proposed in \cite{zhou2017unsupervised,zhang2018activestereonet}.

The key idea in the self-supervised training step is to find disparity maps $D^\text{L/R}_\mathrm{est}$ and validity maps $V^\text{L/R}_\mathrm{est}$ that provide the optimal reconstruction of the stereo images $J^\text{L/R}$ by warping the other images $J^\text{L/R}$ with the disparity $D^\text{L/R}_\mathrm{est}$ in consideration of the validity $V^\text{L/R}_\mathrm{est}$.
The validity maps are defined as the opposite of the occlusion maps $V^\text{L/R}_\mathrm{est} = 1 - O^\text{L/R}_\mathrm{est}$.
In addition to the reconstruction network described in the previous section, we introduce a validation network that estimates the validation maps.
$V^\text{L/R}_{\mathrm{est}}$ to account for occlusion.
For the loss functions, $\mathcal{L}_\text{u}$ encourages the network to estimate disparity maps that reconstruct one stereo view from the other view through disparity warping.
$\mathcal{L}_\text{v}$ is the regularization loss for the validity masks $V^\text{L/R}_{\mathrm{est}}$~\cite{zhang2018activestereonet,riegler2019connecting}.
$\mathcal{L}_\text{d}$ is the disparity smoothness loss.
We train the network parameters of the trinocular reconstruction network and the validation network on the captured stereo images and the illumination image of the prototype.
At the inference time, we mask out the disparity estimates of pixels with low validity.
For further details, refer to the Supplemental Document.
\begin{figure}[t]
\vspace{-5mm}
  \centering
  \includegraphics[width=\linewidth]{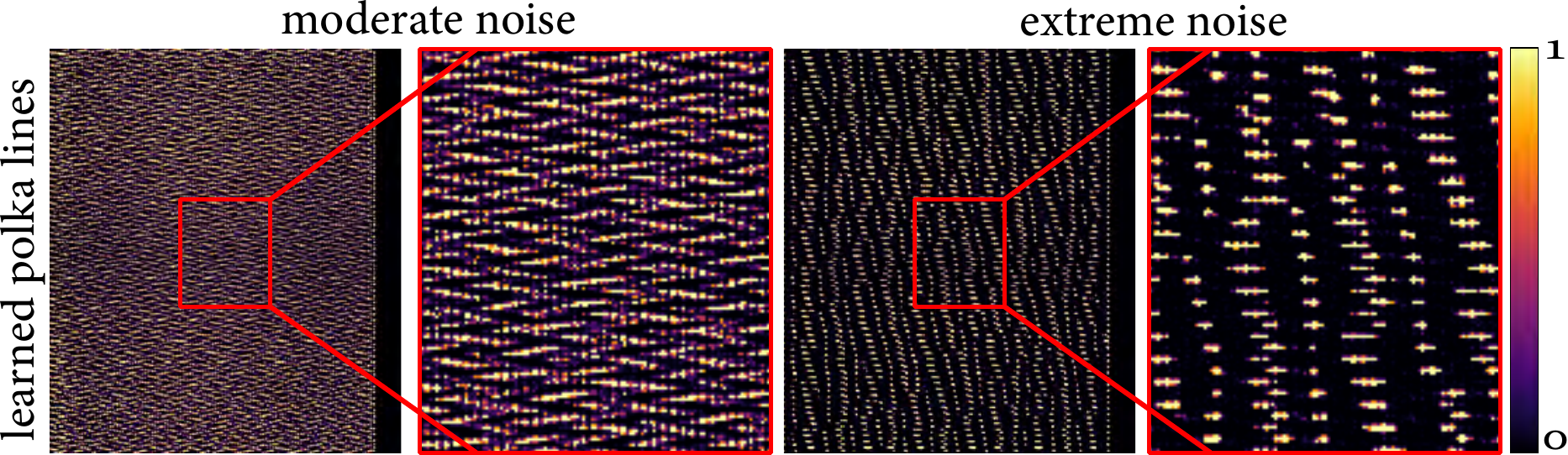}
  \caption{\label{fig:noise_adaptive}
  Optimized illumination for different noise levels. For scenarios with strong ambient light, leading to low illumination contrast, the illumination pattern is optimized to have higher-intensity sparse dots than the moderate noise environment.
  }
  \vspace{-3mm}
\end{figure}
\begin{figure}[t]
  \centering
  \includegraphics[width=0.95\linewidth]{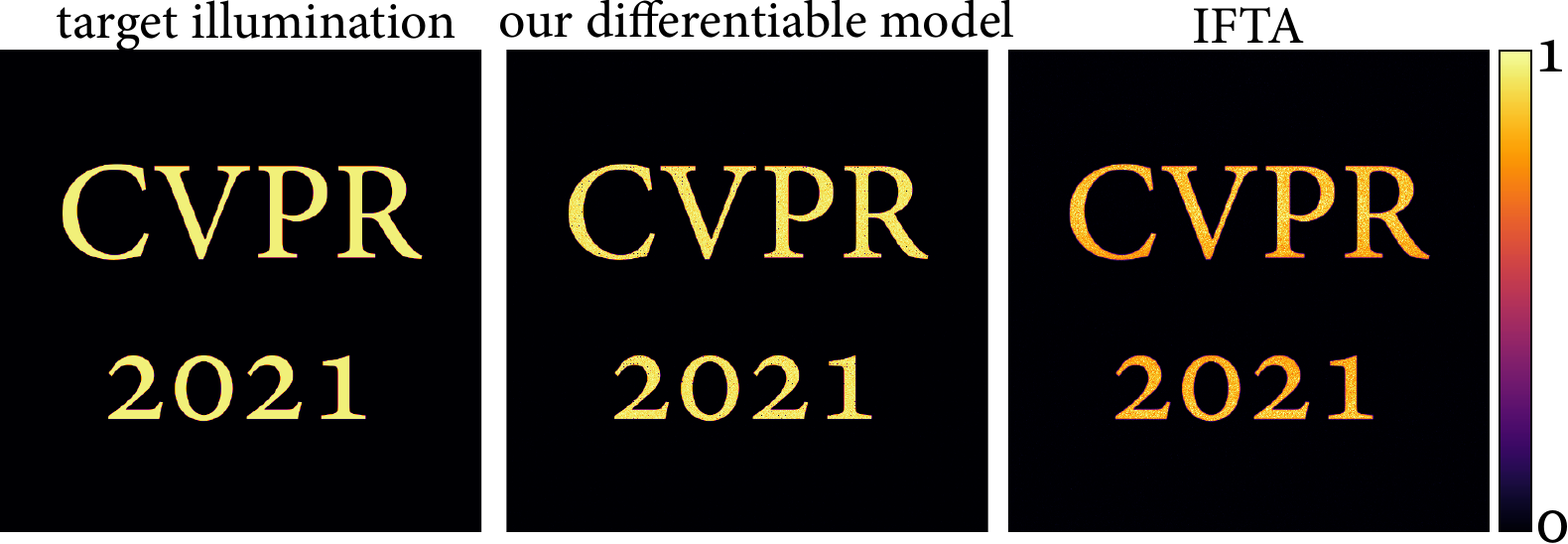}
  \caption{\label{fig:comparison_IFTA}
  The proposed differentiable image formation can be used for designing a DOE that produces the desired illumination pattern. Our method improves on state-of-the-art iterative FFT methods~\cite{du2016design} while allowing for design flexibility, see text.
  }
  \vspace{-5mm}
\end{figure}

\begin{figure*}[t]
\vspace{-5mm}
  \centering
  \includegraphics[width=\linewidth]{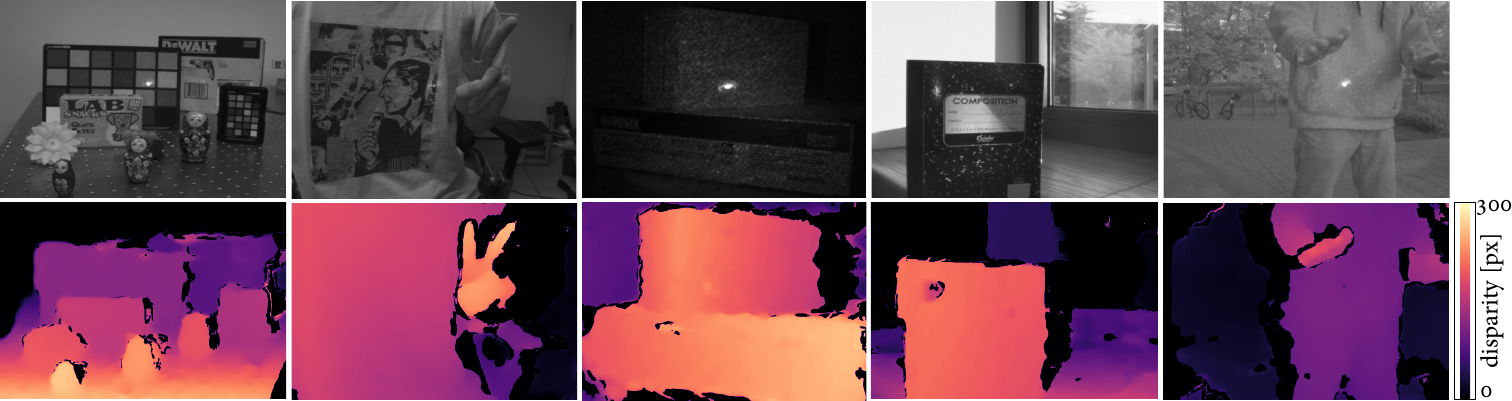}
  \caption{\label{fig:real_qualitative}
  The described system acquires accurate disparity for challenging scenes.
  We show here examples containing complex objects including textureless surface under diverse environments from indoor illumination to outdoor sunlight.
  \vspace{-3mm}
  }
\end{figure*}
\begin{figure}[t]
  \vspace{-5mm}
  \centering
  \includegraphics[width=\linewidth]{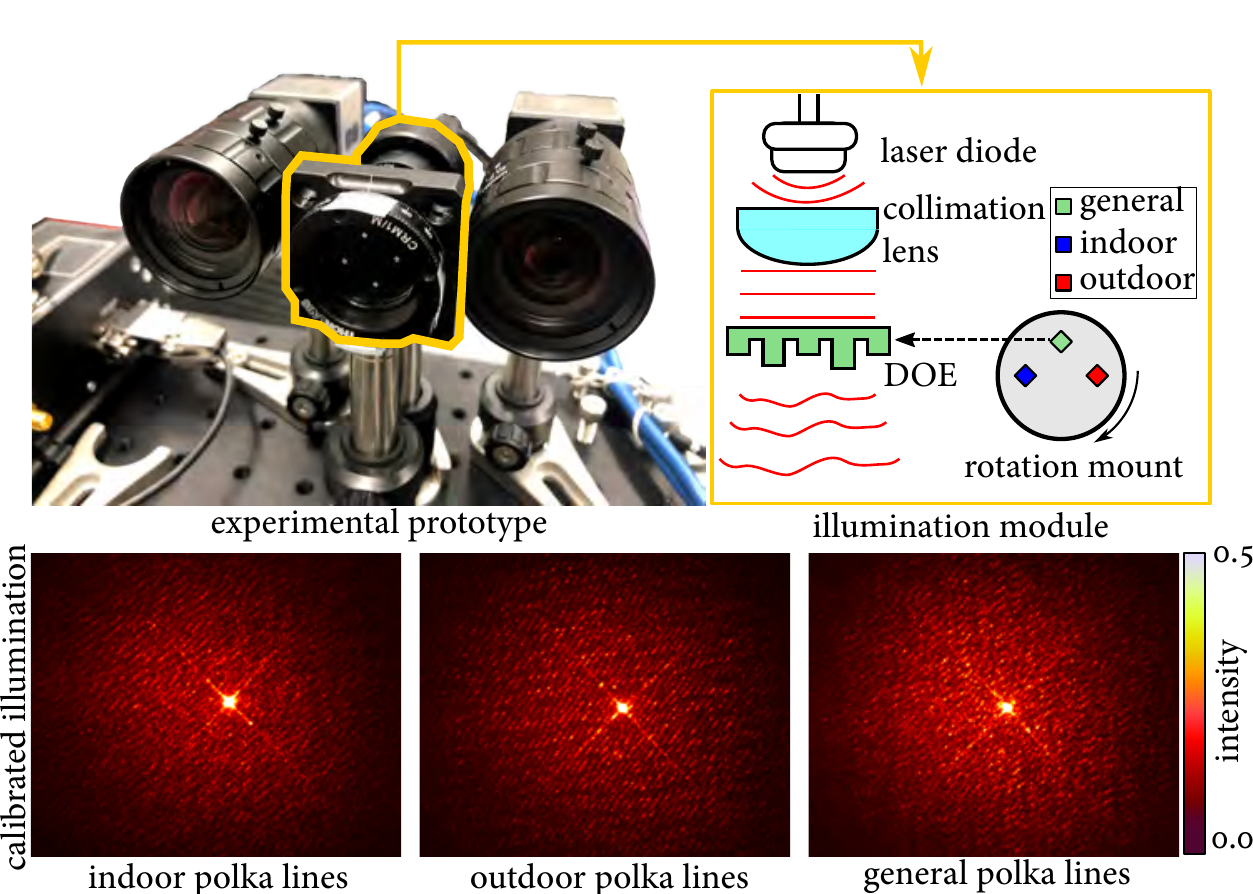}
  \caption{\label{fig:hardware}%
  The proposed prototype system consists of stereo NIR cameras and an illumination module, where laser light is collimated and modulated by a DOE. We fabricated three DOEs designed for generic, indoor, and outdoor environments that can be switched by a rotational mount. Calibrated illumination images closely resemble our simulation; a dense low-intensity dot pattern for the indoor, a sparse high-intensity dot pattern for the outdoor, a dense varying-intensity dot pattern for the generic environment.
  \vspace{-5mm}
  }
\end{figure}

\section{Analysis}
\label{sec:sim_results}
Before introducing our experimental prototype system, we first evaluate the proposed end-to-end framework using synthetic data.

\vspace{0.5em}\noindent\textbf{Polka Lines Illumination Pattern.}\hspace{0.1em}
We evaluate the effectiveness of our learned illumination, the Polka Lines pattern, by comparing to heuristically-designed patterns: the pseudo-random dot and the regularly spaced dot~\cite{intelD415}.
For a fair comparison, we use our trinocular network architecture for all patterns and finetune the reconstruction network for each individual illumination pattern.
The experiments in Figure~\ref{fig:illumination} validate that the proposed Polka Lines pattern outperforms the conventional patterns in indoor environments.
For these synthetic experiments, we ensure that equal illumination power is used for all illumination patterns. We refer to the Supplemental Document for analysis in outdoor environments.
The proposed Polka Lines design is the result of the proposed optimization method. We can interpret the performance of this pattern by analyzing the structure of the Polka Lines patterns compared to heuristic patterns. First, each dot in a line of dots has varying intensity levels, in contrast to the constant-intensity heuristic patterns. We attribute the improved performance in large dynamic ranges to these varying dot intensities. Second, the orientations of Polka Lines are locally varying, which is a discriminative feature for correspondence matching.
We refer to the Supplemental Document for further discussion.

\vspace{0.5em}\noindent\textbf{Trinocular Reconstruction Ablation Study.}\hspace{0.1em}
We validate our trinocular reconstruction method by comparing it to binocular methods such as Zhang et al.\cite{zhang2000flexible}.
We build a baseline model that ingests only binocular inputs of stereo camera images by removing the illumination feature extractor.
Figure~\ref{fig:trinocular} shows that the binocular reconstruction method struggles, especially in occluded regions, where the proposed trinocular approach provides stable estimates.

\vspace{0.5em}\noindent\textbf{Environment-specific Illumination Design.}\hspace{0.1em}
Our end-to-end learning method readily facilitates the design of illumination patterns tailored to specific environments by changing the environment parameters in Equation~\eqref{eq:image_formation} and solving Equation~\eqref{eq:opt}.
We vary the ambient power $\alpha$ and the laser power $\beta$ to simulate indoor, outdoor, and hybrid ``generic'' environments\footnote{We vary the parameter values depending on the environments: indoor ($\alpha=0.0, \beta=1.5$), outdoor ($\alpha=0.5, \beta=0.2$), generic ($\alpha\in[0,0.5], \beta\in[0.2,1.5]$)}. Figure~\ref{fig:ambient} demonstrates that the illumination pattern becomes dense with low-intensity dots in the indoor case for dense correspondence, whereas the outdoor environment promotes a sparse pattern with high-intensity dots that stand out from the ambient light.
In the generic environment, we obtain ``Polka Lines'' with varying intensities from low to high.
We also evaluate the proposed method for two different noise levels, e.g., under strong ambient illumination, using the standard deviation values of 0.02 and 0.6 for the Gaussian noise term $\eta$.
Figure~\ref{fig:noise_adaptive} shows that the illumination pattern becomes sparse with high intensity dotted lines for the severe noise.
\begin{figure}[t]
  \centering
  \includegraphics[width=\linewidth]{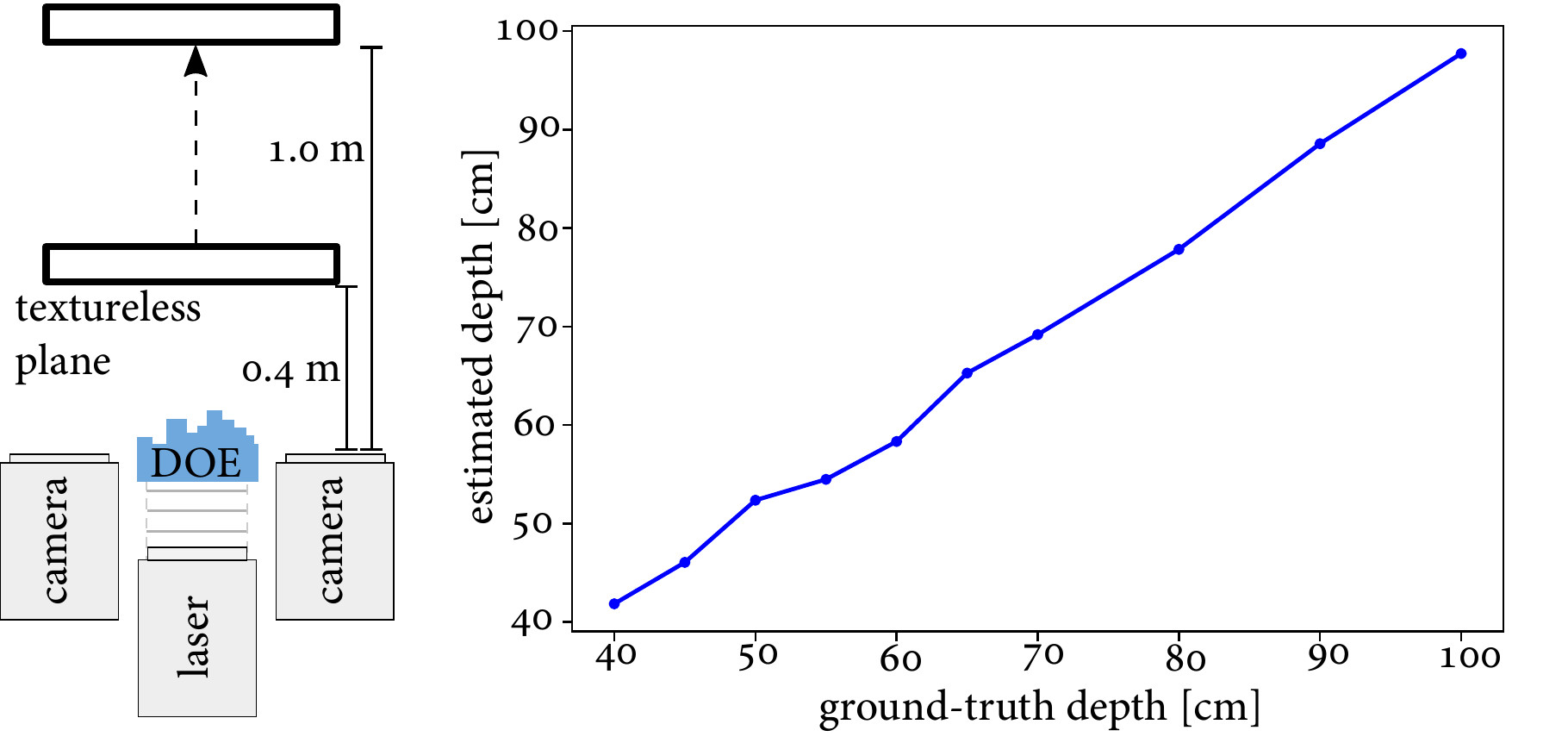}
  \caption{\label{fig:real_quantitative}
  The experimental prototype accurately reconstructs the depth of a textureless plane at distances from 0.4\,m to 1.0\,m.
  }
  \vspace{-5mm}
\end{figure}

\vspace{0.5em}\noindent\textbf{DOE Phase Profile Design.}\hspace{0.1em}
\label{sec:doe_pattern_matching}
We can repurpose the proposed method to design a DOE that produces a target far-field illumination pattern when illuminated by a collimated beam. Designing DOEs for structured illumination has applications beyond active stereo, including anti-fraud protection, projection marking, and surface inspection~\cite{turunen1998diffractive}.
Figure~\ref{fig:comparison_IFTA} shows that we obtain reconstruction quality comparable to state-of-the-art iterative FFT methods~\cite{du2016design}.
One benefit of using our framework for DOE design is its flexibility.
For example, any additional phase-changing optical element can readily be incorporated into the image formation model.
Also, additional loss functions can be imposed, e.g., enforcing smoothness of the DOE to reduce potential fabrication inaccuracies.
We refer to the Supplemental Document for the optimization details.

\begin{figure}[t]
  \centering
  \includegraphics[width=\linewidth]{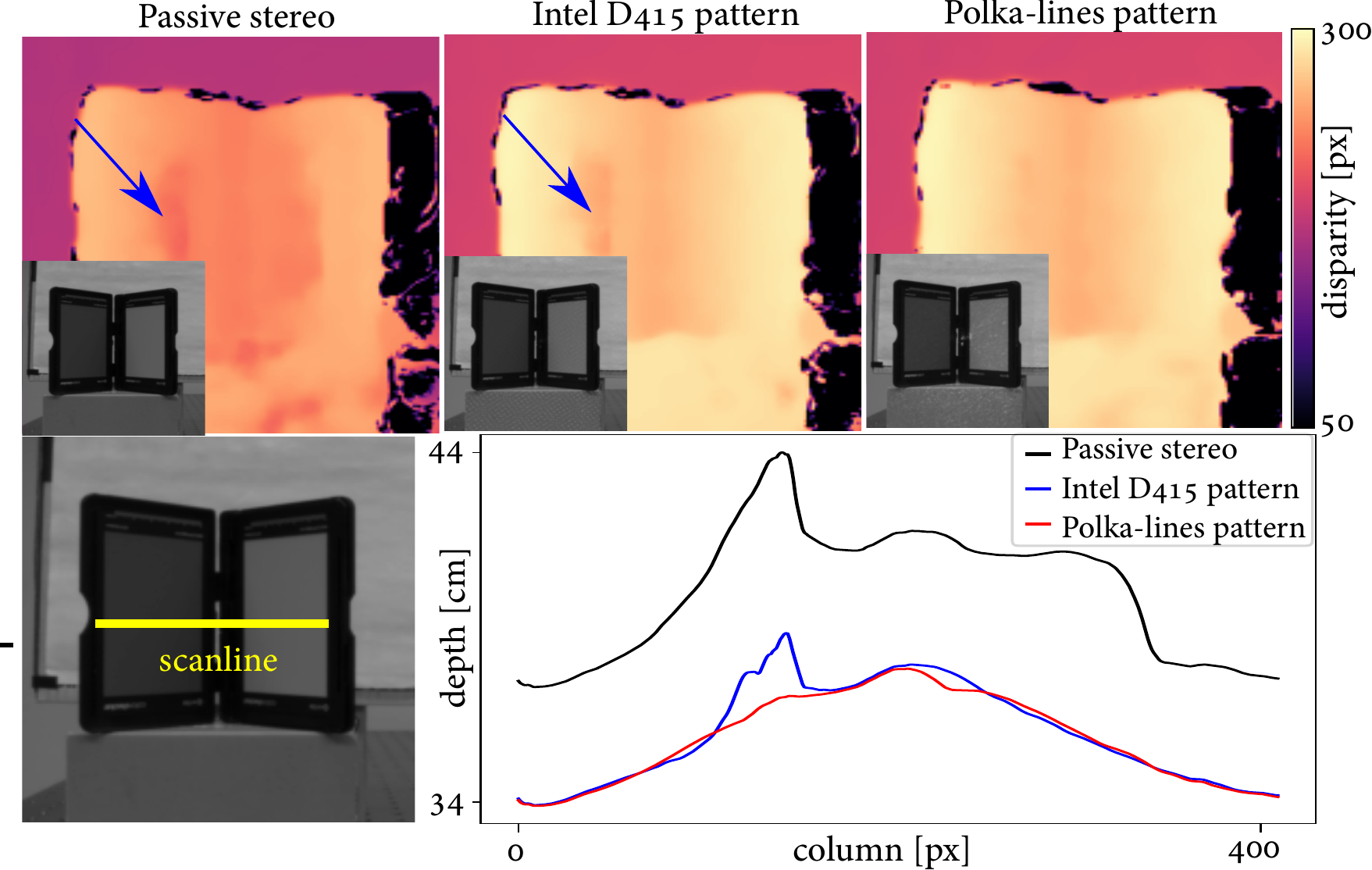}
  \caption{\label{fig:intel_comparison}
  The learned illumination pattern with varying-intensity dots outperforms passive stereo and the commercial hand-engineered pattern (Intel RealSense D415) for high dynamic range scene conditions.
  Blue arrows indicate estimation artifacts.
  We capture a V-shaped reflectance target (x-rite Pro Photo Kit).
  }
  \vspace{-6mm}
\end{figure}

\section{Experimental Prototype Results}
\label{sec:real_results}
\vspace{0.em}\noindent\textbf{Experimental Prototype.}\hspace{0.1em}
Figure~\ref{fig:hardware} shows our experimental prototype along with captures of the proposed Polka Lines illumination pattern variants.
We implement the proposed system with two NIR cameras (Edmund Optics 37-327) equipped with the objective lenses of $\SI{6}{mm}$ focal length (Edmund Optics 67-709).
The pixel pitch of the cameras is $\SI{5.3}{\micro m}$, and the stereo baseline is $\SI{55}{mm}$.
We employ a NIR laser with a center wavelength 850\,nm, and beam diameter of 1\,mm. We use a laser diode (Thorlabs L850P200), a laser diode socket (Thorlabs S7060R), a collimation lens (Thorlabs LT200P-B), and a laser driver (Thorlabs KLD101).
We fabricate the optimized DOE with a 16-level photolithography process. 
For fabrication details, we refer to the Supplemental Document. 
The illumination pattern from the fabricated DOE exhibits undiffracted zeroth-order components that are superposed with the diffracted pattern. While commercial mass-market lithography is highly optimized, our small-batch manual lithography did not meet the same fabrication accuracy. Although the fabrication accuracy is below commercial DOEs with high diffraction efficiency, the measured illumination patterns match their synthetic counterparts.%

\vspace{0.3em}\noindent\textbf{Depth Reconstruction.}\hspace{0.1em}
We measure the depth accuracy of our prototype system by capturing planar textureless objects at known distances as shown in Figure~\ref{fig:real_quantitative}.
The estimated depth using the Polka Lines pattern closely matches the ground truth, with a mean absolute error of 1.4\,cm in the range from 0.4\,m to 1\,m.
We demonstrate qualitative results on diverse real-world scenes in Figure~\ref{fig:real_qualitative}, which includes complex objects, dynamic hand movement, textureless objects without ambient light, objects in sunlight, and moving person in dynamic outdoor environments. We showcase video-rate depth imaging in the Supplemental Video.

\begin{figure}[t]
  \centering
  \includegraphics[width=\linewidth]{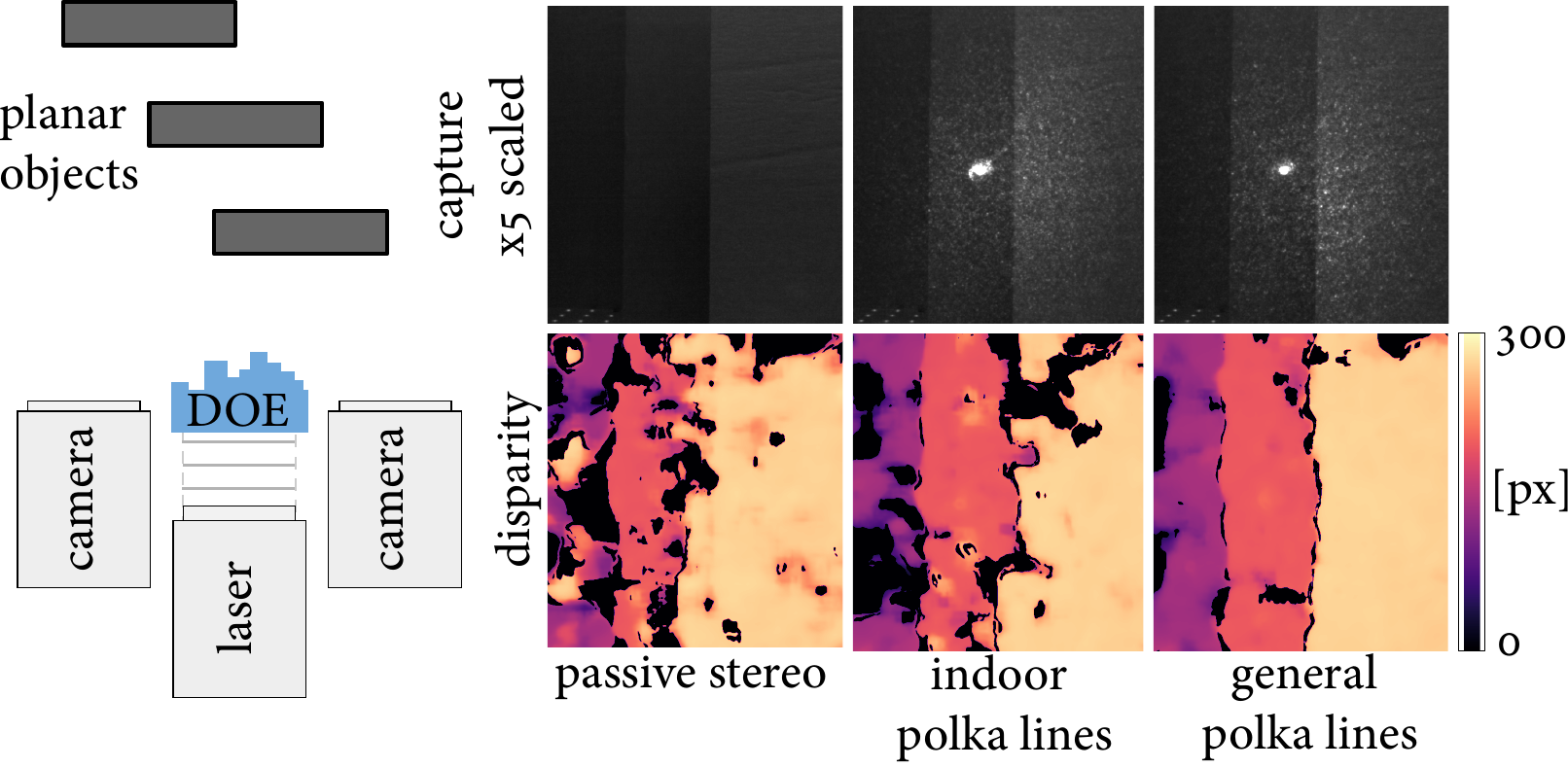}
  \caption{\label{fig:real_doe_compare}
  We capture a scene with low-reflectance planar objects. While passive stereo suffers at the textureless surface, the proposed learned illumination enables effective depth reconstruction. The DOE learned for the generic environment contains a wider range of pattern intensities than the DOE learned for indoor scenes, enabling better depth estimation for these objects.
  }
  \vspace{-5mm}
\end{figure}

\vspace{0.3em}\noindent\textbf{Comparison.}\hspace{0.1em}
We compare our learned Polka Lines pattern with the commercial Intel RealSense D415 pattern in Figure~\ref{fig:intel_comparison}.
The average illumination intensity of the Intel pattern is adjusted to match that of the proposed system via radiometric calibration using an integrating sphere (Thorlabs S142C).
Figure~\ref{fig:intel_comparison} shows that our intensity-varying pattern is more robust to high dynamic range scenes than the Intel pattern, thanks to denser Polka dot patterns with a larger dynamic range.
We note that the Intel pattern is of high fabrication quality and does not exhibit a severe zeroth-order component (as does our fabricated DOE).
We validate our learned Polka Line variants for generic environments and indoor environments in Figure~\ref{fig:real_doe_compare}.
The generic variant features a wide intensity range of dots, resulting in accurate reconstruction for low-reflectance objects.

\section{Conclusion}
\label{sec:conclusion}
We introduce a method for learning an active stereo camera, including illumination, capture, and depth reconstruction. 
Departing from hand-engineered illumination patterns, we learn novel illumination patterns, the Polka Lines patterns, that provide state-of-the-art depth reconstruction and insights on the function of structured illumination patterns under various imaging conditions.
To realize this approach, we introduce a hybrid image formation model that exploits both wave optics and geometric optics for efficient end-to-end optimization, and a trinocular reconstruction network that exploits the trinocular depth cues of active stereo systems.
The proposed method allows us to design environment-specific structured Polka Line patterns tailored to the camera and scene statistics.
We validate the effectiveness of our approach with comprehensive simulations and with an experimental prototype, outperforming conventional hand-crafted patterns across all tested scenarios. In the future, combined with a spatial light modulator, the proposed method may not only allow for ambient illumination specific patterns, but also semantically driven dynamic illumination patterns that adaptively increase depth accuracy. 
\section*{Acknowledgements}
\label{sec:acknowledgements}
The authors are grateful to Ethan Tseng and Derek Nowrouzezahrai for fruitful discussions. Felix Heide was supported by an NSF CAREER Award (2047359) and a Sony Young Faculty Award. 

{\small
\bibliographystyle{ieee_fullname}
\bibliography{egbib}
}

\end{document}